\documentclass{article}

\usepackage{microtype}
\usepackage{graphicx}
\usepackage{subfigure}
\usepackage{booktabs} 
\usepackage{amsmath,amssymb}

\usepackage{hyperref}

\usepackage[accepted]{icml2018}

\icmltitlerunning{Continual Learning in Deep Neural Network by Using a Kalman Optimiser}

\begin{document}

\twocolumn[
\icmltitle{Continual Learning in Deep Neural Network by Using a Kalman Optimiser}



\icmlsetsymbol{equal}{*}

\begin{icmlauthorlist}
\icmlauthor{Honglin Li}{surrey}
\icmlauthor{Shirin Enshaeifar}{surrey}
\icmlauthor{Frieder Ganz}{adobe}
\icmlauthor{Payam Barnaghi}{surrey}
\end{icmlauthorlist}

\icmlaffiliation{surrey}{Centre for Vision, Speech and Signal Process (CVSSP), University of Surrey, United Kingdom}
\icmlaffiliation{adobe}{Adobe, Germany}
\icmlcorrespondingauthor{Honglin Li}{h.li@surrey.ac.uk}

\icmlkeywords{Catastrophic forgetting, Kalman Optimizer}
\vskip 0.3in
]



\printAffiliationsAndNotice{}  

\begin{abstract}
Learning and adapting to new distributions or learning new tasks sequentially without forgetting the previously learned knowledge is a challenging phenomenon in continual learning models. Most of the conventional deep learning models are not capable of learning new tasks sequentially in one model without forgetting the previously learned ones. We address this issue by using a Kalman Optimiser. The Kalman Optimiser divides the neural network into two parts: the long-term and short-term memory units. The long-term memory unit is used to remember the learned tasks and the short-term memory unit is to adapt to the new task. We have evaluated our method on MNIST, CIFAR10, CIFAR100 datasets and compare our results with state-of-the-art baseline models. The results show that our approach enables the model to continually learn and adapt to the new changes without forgetting the previously learned tasks. 
\end{abstract}
\section{Introduction}
\begin{figure}[t]
    \centering
        \subfigure[]{\includegraphics[width=0.45\linewidth,height=80pt]{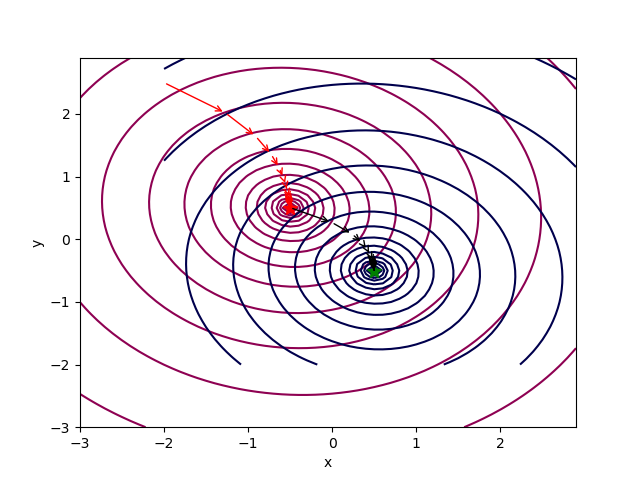}\label{fig:gradient}}
        \subfigure[]{\includegraphics[width=0.45\linewidth,height=80pt]{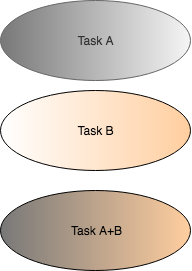}\label{fig:intro_merge}}
        \caption{(a) Model find different solutions for different tasks (b) Different part of the parameters adapts to different tasks; e.g. if the important parameters of task A are on the left, we then let the parameters on the right side adapt to task B}
\end{figure}

Conventional deep learning models have achieved significant successes in a variety of fields including computer vision and speech recognition. However, most of the dominant models have to be trained with all the expected tasks or variations in the data at the same time. Otherwise, they tend to forget the learned knowledge when they switch between different tasks and various datasets in a periodically changing environment. The problem that the model forgets how to perform on the previously learned tasks is often referred to as catastrophic forgetting \cite{mccloskey1989catastrophic,goodfellow2013empirical}. This issue often occurs when a model adjusts its parameters to cater for new tasks and when the newly set parameters are not suitable anymore to provide accurate results to the previously learned tasks when they occur again. The parameters may significantly change when the training task is very different from previously learned ones. For example, in Figure \ref{fig:gradient}, training a neural network is aimed at finding an ideal solution (i.e. the red circle in Figure \ref{fig:gradient}) for the training data. When the model continually learns with a new set of data, the neural network will find another ideal solution (i.e. the green circle in Figure \ref{fig:gradient}). The new solution could be very different from the previous one. Consequently, the performance of the neural network on the previously learned task(s) would decrease. In other words, the neural network forgets how to perform on the old dataset.
This learning process is very different from the biological learning process which can acquire new knowledge sequentially. In real-world scenarios, we cannot ensure that our training data is the most representative set and it may not cover all the tasks in advance. Possible solutions to address this issue include training a new model for each new task, re-training the model when the previously learned tasks reappear or are required again and storing all the training data and frequently training the model based on the whole datasets. Storing all the training data is very inefficient and requires high resource. Training the model again and again for re-appearing tasks and goals is very inefficient and computationally costly. Our goal is to develop a continual learning model in a way that the model can learn new tasks without forgetting the previously learned ones. 

To solve this problem, ensemble learning is among the solutions that are proposed in the existing work \cite{wozniak2014survey,polikar2001learn++,Dai}. The fundamental idea in ensemble learning is to build a network for each task or learning target. As a consequence, $n$ networks would be created ($n$ is the number of the tasks to be learned). This method is not always a desirable solution because of high memory requirements and complexity \cite{kemker2017measuring}. \cite{fernando2017pathnet} propose PathNet which is based on ensemble learning but offers less complexity. In PathNet, the learned networks can contribute to train the model while learning new tasks. Different from ensemble learning, the dual-memory-based learning approach interleaves the new training data with the learned samples offers less complexity. \cite{lopez2017gradient} propose Gradient Episodic Memory (GEM) to deal with the forgetting issue. GEM stores a subset of the samples that are used in the new training process. While training on new tasks, the losses on the old samples are only allowed to decrease. Other recent developments in this direction reduce the memory requirements of old knowledge by leveraging a pseudo-rehearsal technique; e.g.,  \cite{robins1995catastrophic} construct probabilistic models to generate training samples (based on what has been seen in the past) to reduce the memory resources required for storing and reloading large volumes of old samples. \cite{shin2017continual} propose an architecture consisting of deep generative models to generate training samples and task solvers to perform a classification task. Similarly, \cite{kamra2017deep} use a variational autoencoder to generate the training samples. \cite{hinton1987using} propose a dual-memory system in which each synaptic connection has two weights: a plastic weight with slow change rate for long-term knowledge and a fast-changing weight for short-term knowledge. Another solution called Regularisation-based method does not need to produce multiple models or store the trained samples.  \cite{kirkpatrick2017overcoming} propose the Elastic Weight Consolidation (EWC) to overcome the forgetting problem. EWC pulls back the changes of parameters when the models require to carry out a previously learned task. \cite{huszar2017quadratic} extended the EWC to a continuous technique that learns the new tasks recursively. Similarly, \cite{zenke2017continual} attempt to address the forgetting problem by avoiding to change the influential parameters that contribute to carrying out the previously learned tasks.

Although there are various works to address the forgetting problem in continual learning scenarios, few of the existing work proposes an optimiser or leverages the uncertainty information in the neural networks to resolve the catastrophic forgetting problem. In this paper, we propose a Kalman-Filter-based optimiser. Some existing works that combine the Kalman Filter with neural networks. For instance, \cite{ollivier2017online} demonstrate an extended Kalman Filter to estimate the parameters similar to an online stochastic natural gradient descent. \cite{trebaticky2005recurrent} use a Kalman Filter to train a Recurrent neural network. \cite{haarnoja2016backprop} combine the Kalman Filter with a feed-forward Convolutional neural network. \cite{wu2012extended} propose an Extended and Unscented Kalman Filter training algorithm for training feed-forward neural networks. Different from previous work, our key idea is to find an optimal solution for all the tasks by letting different part of the parameters adapt to different tasks, see Figure \ref{fig:intro_merge}. More specifically, we obtain the important parameters in the previously learned tasks and group them in long-term memory units. The other parameters are grouped into short-term memory units. Different from the classical Long Short Term Memory networks (LSTM) and \cite{hinton1987using}, we only use a single neural network layer, and the parameters that are involved in each of the units are dynamically decided at the end of each training process. We use a Kalman Filter to restrict the changes of the parameter in the long-term memory units. Then the short-term memory units adapt to the new tasks.

\section{Methodology}
\subsection{Gradient as an Uncertainty Measure}
We consider the weights as the indeterminate values instead of the deterministic values. Different from the Bayesian neural networks \cite{blundell2015weight}, our goal is to track the changes of the values and uncertainties of the parameters and then adjust them. To determine the uncertainty, we use the gradient as a measure since it reflects how the model is uncertain with the current parameters given the current data. From the gradient descent point of view, if the parameter is far from the optimal value, the gradient of the parameter is larger. In other words, the model would be highly uncertain with this parameter given the training data. For example, in Figure \ref{fig:gradient}, the training parameters take a larger step at the beginning of training since they are far from the optimal value. As the training continues, the step becomes smaller and smaller. Later on, and when this model is trained on a new task, this process is repeated. 
\subsection{Kalman Optimizer}
Based on the values of the weights and the uncertainty measure, we restrict the changes of the weights to let the new value have lower uncertainty by using a Kalman Filter \cite{rhudy2017kalman,enshaeifar2016regularised}. At the end of the first training, we consider the uncertainty and the optimal solution of our first task as our prior knowledge. During the training for the second task, we track the changes in the weights and uncertainty. We then use a Kalman Filter to adjust the weights based on our prior knowledge. The predicted values of the Kalman Filter would be close to the values that will result in lower uncertainty. 
Given the initial information $\theta_0$ and $P_0$, where $\theta_0$ is the set of parameters in the pre-trained model and $P_0$ is the uncertainty of model on the previously trained dataset or task, $\xi$ is a very small hyper-parameter in the case that the denominator is zero. While training on the new dataset or for a new task, according to the gradient descent process and mini-batch algorithm, at batch $k$, we can predict the value $\hat{\theta}_{k|k-1}$ and obtain an uncertainty measure $R_k$ which refers to the gradient of the parameter $\theta_{k-1}$ on the $k_{th}$ batch data. The Kalman Gain $K_k$, the optimal value $\hat{\theta}_{k}$ and the uncertainty $P_k$ can be calculated according to the following equations:
\begin{subequations}
\label{eq:kal_update}
    \begin{align}
        \hat{\theta}_{k|k-1} &= \theta_{k-1} - learning\_rate \times Gradients \\
        K_k &= \frac{P_{k-1} } { (P_{k-1} + R_k + \xi) } \\
        \hat{\theta}_{k} &= \theta_{k-1} + K_k (\hat{\theta}_{k|k-1} - \theta_{k-1}) \\
        P_k &= (I - K_k) P_{k-1}
    \end{align}
\end{subequations}
Since $R_{k}$ is the gradient of the model on the new dataset or task, it would be relatively higher than $P_{k-1}$, especially in the beginning of the training process. This means $P_{k-1}$ would decrease more rapidly compared with $R_{k}$. Hence, the predicted values would be close to the previous optimal solution.

However, the model cannot learn the new task very well, if all the weights are close to the previous optimal solution. To address this issue, we let the Kalman Optimiser identify which part of the parameters should adjust to remember the learned knowledge and which part of the parameters should have less influence on learning the new task. To achieve this goal, we find the important parameters to the previously learned tasks and group them as the long-term memory, and group the rest of the parameters as the short-term memory. To find the important parameters, we use the Fisher Information matrix. 
 The Fisher Information matrix is the approximation of the second order derivatives of the loss near a minimal point. We assume that the covariance matrix of the posterior distribution for a trained task is diagonal and obtain the Fisher Information by using Equation (\ref{eq:fishercalculate}), where $\theta$ represents the parameters, $D$ is the training dataset, $F$ is the Fisher Information matrix.
\begin{equation}
    \label{eq:fishercalculate}
    F(\theta) = \mathbb{E}_{\theta}[\frac{\partial}{\partial \theta}\log(D,\theta)]^2
\end{equation}
We further normalise the Fisher Information by using Equation (\ref{eq:fisher}) to obtain the rate of importance for the parameters:
\begin{equation}
    \label{eq:fisher}
    F(\theta)^* = \frac{F(\theta)}{\max F(\theta)} 
\end{equation}
We then use the normalised Fisher Information to divide the units into two different categories (i.e. long-term and short-term memory units). We set a threshold to distinguish the boundary to choose the significant parameters that are involved in long-term memory. The final update procedure is shown in Equation (\ref{eq:kal_improvement}) where $\alpha$ is a pre-defined threshold. Hence the Kalman Optimise can identify what parameters to adjust instead of adjusting all of them.
\begin{subequations}
\label{eq:kal_improvement}
\begin{align}
            &F^* = 
    \begin{cases}
    F^*, \mbox{if }F^* < \alpha \\
    1, otherwise
    \end{cases} \\
    \begin{split}
                \hat{\theta}_{k} &= \theta_{k-1} + (\hat{\theta}_{k|k-1} - \theta_{k-1}) * K_k * F^*\\
     &+ (\hat{\theta}_{k|k-1} - \theta_{k-1}) * (1 - F^*) 
    \end{split}\\
    P_k &= (I - K_k * F^*) P_{k-1} 
    \end{align}
\end{subequations}
While learning several tasks, \cite{kirkpatrick2017overcoming} use multiple Fisher Information matrices to apply the constraints on parameters. This requires high computation and increases the complexity. Computing and storing large number of Fisher Information can also quickly become intractable. In our proposed method, only one Fisher Information matrix is needed. In the Kalman Optimizer, only the larger value among the different Fisher Information Matrix will be stored in the memory. We update the normalised Fisher Information recursively and this addresses the complexity and scalability issues related to the computation and storing large number of Fisher Information. 

The last step is to update the uncertainty information. Up to this stage, all the prior knowledge is updated by the Kalman Filter. At this stage, we update our prior knowledge at the end of each training, in case the Kalman Filter is converged into a constant. Mathematically, $P$, which is the uncertainty of the model on the learned tasks, could be a minimal value at the end of the training process. If the uncertainty is not updated, the value predicted by the Kalman Optimiser will be close to a constant. Hence, after training task $k$, if the parameter is more important to task $k$ than other tasks, we then update the uncertainty of this parameter given the data in task $k$. The other uncertainties remain the same since we cannot access to the previous dataset anymore in our experiments. Furthermore, these small uncertainties are very helpful to preserve the learned knowledge.

\section{Experiments}
We evaluate our method by sequentially learning the disjoint MNIST \cite{lecun2010mnist}, CIFAR10 and CIFAR100 datasets \cite{krizhevsky2009cifar}. These experiments are commonly used to evaluate the performance of the methods that address the forgetting problem in continual learning scenarios.
\begin{figure}[t]
\vskip 0.2in
    \centering
    \subfigure[Task 1]{\includegraphics[width=0.45\linewidth]{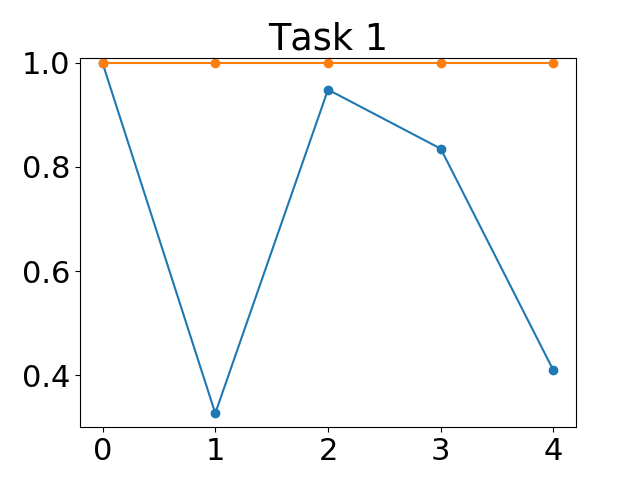}\label{fig:split_mnist_task1}}
     \subfigure[Task 2]{\includegraphics[width=0.45\linewidth]{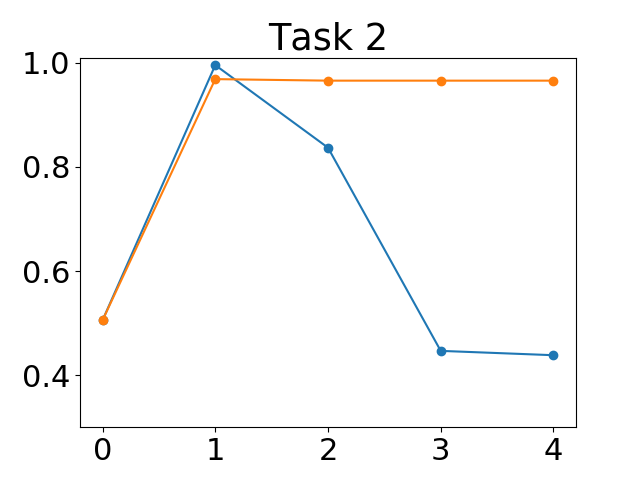}\label{fig:split_mnist_task2}}
    \subfigure[Task 3]{\includegraphics[width=0.45\linewidth]{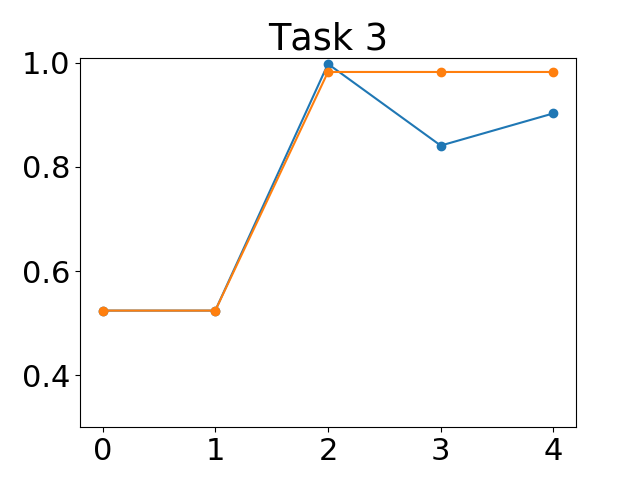}\label{fig:split_mnist_task3}}
    \subfigure[Task 4]{\includegraphics[width=0.45\linewidth]{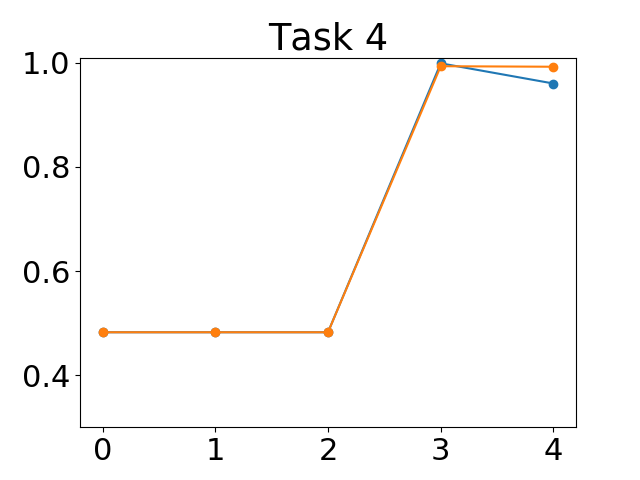}\label{fig:split_mnist_task4}}
    \subfigure[Task 5]{\includegraphics[width=0.45\linewidth]{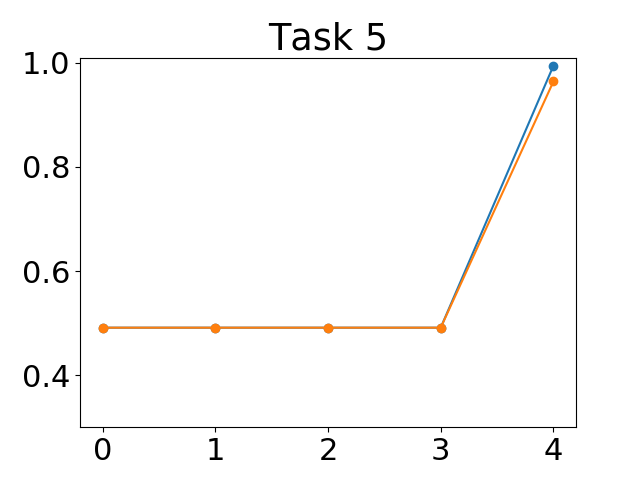}\label{fig:split_mnist_task5}}
    \subfigure[All Task]{\includegraphics[width=0.45\linewidth]{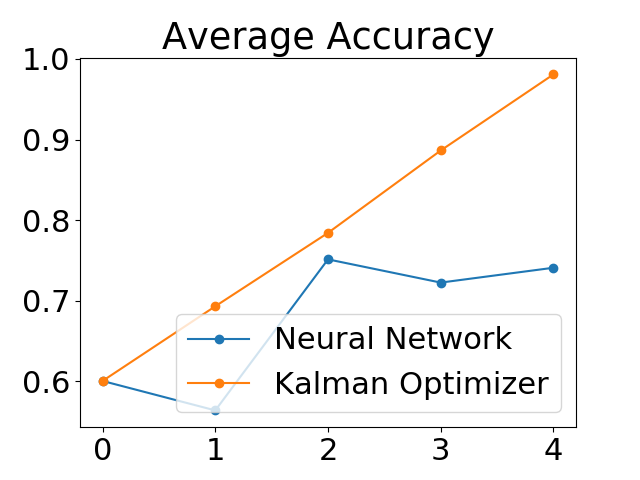}\label{fig:split_mnist_avg}}
    \caption{The evaluation results for the disjoint MNIST}
    \label{fig:split_mnist}
    \end{figure}
\subsection{Disjoint MNIST}
The first experiment is conducted by using the disjoint MNIST dataset. We split the MNIST dataset into 5 subsets of consecutive digits from 0 to 10. We use a shallow neural network containing two hidden layers consisting 256 units. Due to the label distribution changes in different task, we use a multi-head approach which only computes the loss for the digits present in the current task \cite{zenke2017continual}. The results are shown in Figure \ref{fig:split_mnist}.
\begin{figure}[t]
\vskip 0.2in
    \centering
    \subfigure[Task 1]{\includegraphics[width=0.32\linewidth]{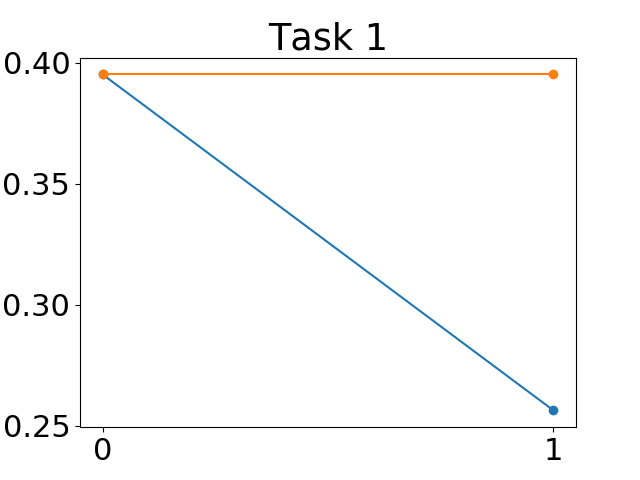}\label{fig:cifar10_task1}}
     \subfigure[Task 2]{\includegraphics[width=0.32\linewidth]{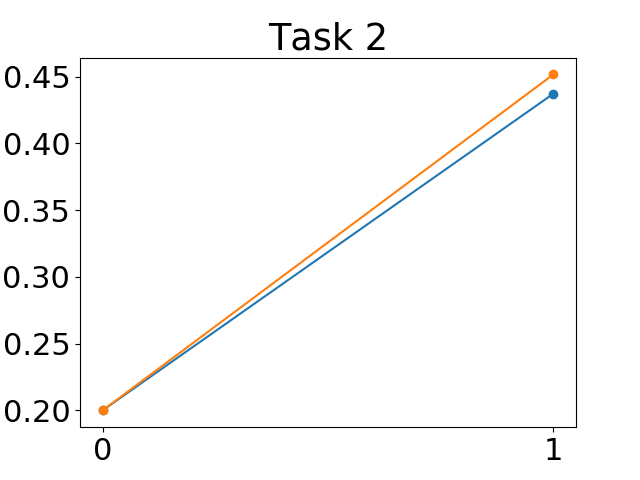}\label{fig:cifar10_task2}}
    \subfigure[All Task]{\includegraphics[width=0.32\linewidth]{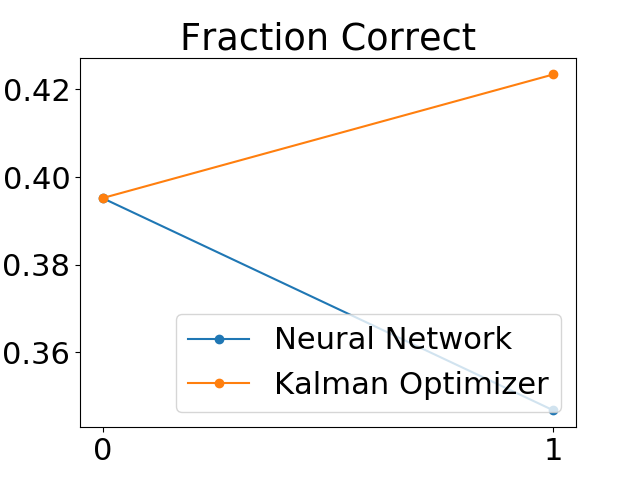}\label{fig:cifar10_task}}
    \caption{The evaluation results for the disjoint CIFAR10}
    \label{fig:split_cifar}
    \end{figure}
We compare the performance of the model with and without the proposed optimiser. To evaluate the performance, we compute the average accuracy of the model on all the tasks. The aim of the training is to have an average accuracy close to 1.0 after training the model on all the tasks. The first two figures in Figure \ref{fig:split_mnist} show that the performance of the neural network drops to a guesstimate level (less than 50\%) after training on all the tasks. The performance of the third task is also decreased. However, the neural network with the proposed optimiser maintains the performance on all the learned tasks. The accuracy of the proposed method on the first two tasks stays close to 1.0, while the average accuracy of the model on all the tasks keeps increasing. 

\subsection{Disjoint CIFAR10 and CIFAR100}
\begin{figure}[t]
    \centering
        \subfigure[CIFAR10]{\includegraphics[width=0.3\linewidth]{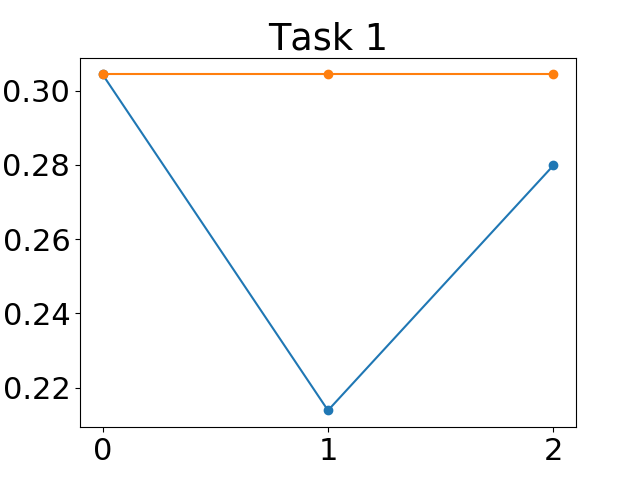}\label{fig:cifar100_d1}}
     \subfigure[CIFAR100- 1]{\includegraphics[width=0.3\linewidth]{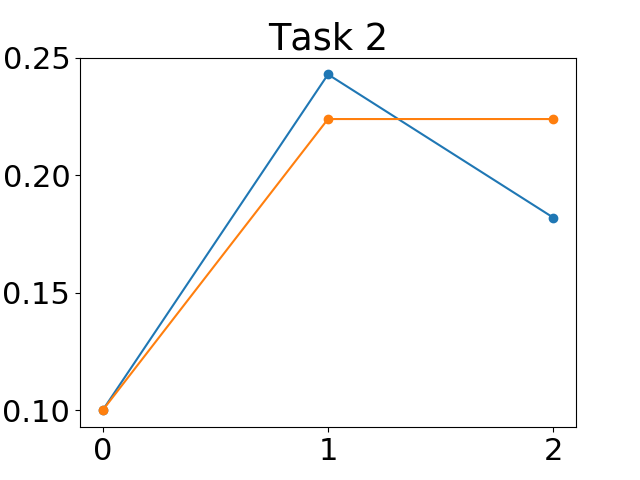}\label{fig:cifar100_d2}}
          \subfigure[CIFAR100- 2]{\includegraphics[width=0.3\linewidth]{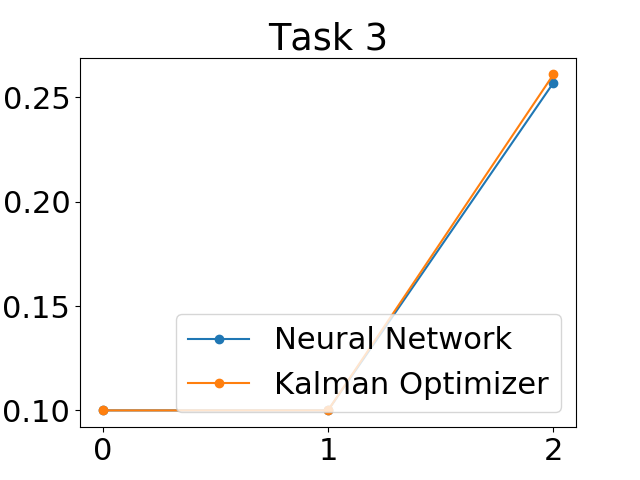}\label{fig:cifar100_d3}}
        \caption{The evaluation results from CIFAR10 to CIFAR100}
        \label{fig:cifar10to100}
\end{figure}

This experiment consists of two parts. In the first part, we evaluate our method based on the disjoint CIFAR10 dataset. We split the CIFAR10 into two subsets of consecutive classes. We use a Convolutional Neural Network containing four convolutional layers, two fully-connected layers and also the multi-head approach. The results are shown in Figure \ref{fig:split_cifar}. As shown in Figure \ref{fig:cifar10_task1}, the performance of the common model on the first task after learning the subsequent tasks and then by revisiting the first task decreases dramatically. However, the proposed method still remembers how to perform on the first task. The fraction accuracy, which is the average accuracy on all previous tasks (Figure \ref{fig:cifar10_task}), of the proposed method keeps increasing while the fraction accuracy of the baseline model decreases steadily. In the second part, we evaluate our method on CIFAR100. The first task is the original CIFAR10 dataset, the second and third tasks are ten different classes from CIFAR100. The other settings are the same. The results are shown in Figure \ref{fig:cifar10to100}. The performance of the model on the learned tasks decreases while the proposed method remembers how to perform the learned tasks. 
\section{Conclusions}
We present a novel optimisation method to learn and adapt to the new tasks without forgetting the previously learned tasks. The key idea is to find an optimal solution by letting different parts of the parameters adapt to different tasks. The proposed method uses the gradients to obtain an uncertainty measure and groups the learning parameters into long-term and short-term memory units. These units define which parameters are restricted to be changed by the Kalman update procedure (i.e. long-term memory). This update procedure adds an adjustment and control mechanism to allow the model to learn new tasks without significantly forgetting the previously learned ones. We evaluate our method based on the disjoint MNIST and CIFAR10 datasets, and also by continually learning from CIFAR10 to CIFAR100 and compared the results to a baseline. The results show that the proposed method can preserve the previously learned knowledge and efficiently learn and adapt to the new tasks.
\section*{Acknowledgments}
This work is partially supported by the EU H2020 IoTCrawler project under contract number: 779852.

\bibliographystyle{icml2018}
\bibliography{ref}

\begin{thebibliography}{24}
\providecommand{\natexlab}[1]{#1}
\providecommand{\url}[1]{\texttt{#1}}
\expandafter\ifx\csname urlstyle\endcsname\relax
  \providecommand{\doi}[1]{doi: #1}\else
  \providecommand{\doi}{doi: \begingroup \urlstyle{rm}\Url}\fi

\bibitem[Blundell et~al.(2015)Blundell, Cornebise, Kavukcuoglu, and
  Wierstra]{blundell2015weight}
Blundell, Charles, Cornebise, Julien, Kavukcuoglu, Koray, and Wierstra, Daan.
\newblock Weight uncertainty in neural networks.
\newblock \emph{arXiv preprint arXiv:1505.05424}, 2015.

\bibitem[Dai et~al.(2007)Dai, Yang, Xue, and Yu]{Dai}
Dai, Wenyuan, Yang, Qiang, Xue, Gui-Rong, and Yu, Yong.
\newblock Boosting for transfer learning.
\newblock In \emph{Proceedings of the 24th International Conference on Machine
  Learning}, ICML '07, pp.\  193--200, New York, NY, USA, 2007. ACM.
\newblock ISBN 978-1-59593-793-3.
\newblock \doi{10.1145/1273496.1273521}.
\newblock URL \url{http://doi.acm.org/10.1145/1273496.1273521}.

\bibitem[Enshaeifar et~al.(2016)Enshaeifar, Spyrou, Sanei, and
  Took]{enshaeifar2016regularised}
Enshaeifar, Shirin, Spyrou, Loukianos, Sanei, Saeid, and Took, Clive~Cheong.
\newblock A regularised eeg informed kalman filtering algorithm.
\newblock \emph{Biomedical Signal Processing and Control}, 25:\penalty0
  196--200, 2016.

\bibitem[Fernando et~al.(2017)Fernando, Banarse, Blundell, Zwols, Ha, Rusu,
  Pritzel, and Wierstra]{fernando2017pathnet}
Fernando, Chrisantha, Banarse, Dylan, Blundell, Charles, Zwols, Yori, Ha,
  David, Rusu, Andrei~A, Pritzel, Alexander, and Wierstra, Daan.
\newblock Pathnet: Evolution channels gradient descent in super neural
  networks.
\newblock \emph{arXiv preprint arXiv:1701.08734}, 2017.

\bibitem[Goodfellow et~al.(2013)Goodfellow, Mirza, Xiao, Courville, and
  Bengio]{goodfellow2013empirical}
Goodfellow, Ian~J, Mirza, Mehdi, Xiao, Da, Courville, Aaron, and Bengio,
  Yoshua.
\newblock An empirical investigation of catastrophic forgetting in
  gradient-based neural networks.
\newblock \emph{arXiv preprint arXiv:1312.6211}, 2013.

\bibitem[Haarnoja et~al.(2016)Haarnoja, Ajay, Levine, and
  Abbeel]{haarnoja2016backprop}
Haarnoja, Tuomas, Ajay, Anurag, Levine, Sergey, and Abbeel, Pieter.
\newblock Backprop kf: Learning discriminative deterministic state estimators.
\newblock In \emph{Advances in Neural Information Processing Systems}, pp.\
  4376--4384, 2016.

\bibitem[Hinton \& Plaut(1987)Hinton and Plaut]{hinton1987using}
Hinton, Geoffrey~E and Plaut, David~C.
\newblock Using fast weights to deblur old memories.
\newblock In \emph{Proceedings of the ninth annual conference of the Cognitive
  Science Society}, pp.\  177--186, 1987.

\bibitem[Husz{\'a}r(2017)]{huszar2017quadratic}
Husz{\'a}r, Ferenc.
\newblock On quadratic penalties in elastic weight consolidation.
\newblock \emph{arXiv preprint arXiv:1712.03847}, 2017.

\bibitem[Kamra et~al.(2017)Kamra, Gupta, and Liu]{kamra2017deep}
Kamra, Nitin, Gupta, Umang, and Liu, Yan.
\newblock Deep generative dual memory network for continual learning.
\newblock \emph{arXiv preprint arXiv:1710.10368}, 2017.

\bibitem[Kemker et~al.(2017)Kemker, McClure, Abitino, Hayes, and
  Kanan]{kemker2017measuring}
Kemker, Ronald, McClure, Marc, Abitino, Angelina, Hayes, Tyler, and Kanan,
  Christopher.
\newblock Measuring catastrophic forgetting in neural networks.
\newblock \emph{arXiv preprint arXiv:1708.02072}, 2017.

\bibitem[Kirkpatrick et~al.(2017)Kirkpatrick, Pascanu, Rabinowitz, Veness,
  Desjardins, Rusu, Milan, Quan, Ramalho, Grabska-Barwinska,
  et~al.]{kirkpatrick2017overcoming}
Kirkpatrick, James, Pascanu, Razvan, Rabinowitz, Neil, Veness, Joel,
  Desjardins, Guillaume, Rusu, Andrei~A, Milan, Kieran, Quan, John, Ramalho,
  Tiago, Grabska-Barwinska, Agnieszka, et~al.
\newblock Overcoming catastrophic forgetting in neural networks.
\newblock \emph{Proceedings of the national academy of sciences}, pp.\
  201611835, 2017.

\bibitem[Krizhevsky et~al.(2009)Krizhevsky, Nair, and
  Hinton]{krizhevsky2009cifar}
Krizhevsky, Alex, Nair, Vinod, and Hinton, Geoffrey.
\newblock Cifar-10 and cifar-100 datasets.
\newblock \emph{URl: https://www. cs. toronto. edu/kriz/cifar. html}, 6, 2009.

\bibitem[LeCun et~al.(2010)LeCun, Cortes, and Burges]{lecun2010mnist}
LeCun, Yann, Cortes, Corinna, and Burges, CJ.
\newblock Mnist handwritten digit database.
\newblock \emph{AT\&T Labs [Online]. Available: http://yann. lecun.
  com/exdb/mnist}, 2, 2010.

\bibitem[Lopez-Paz et~al.(2017)]{lopez2017gradient}
Lopez-Paz, David et~al.
\newblock Gradient episodic memory for continual learning.
\newblock In \emph{Advances in Neural Information Processing Systems}, pp.\
  6467--6476, 2017.

\bibitem[McCloskey \& Cohen(1989)McCloskey and
  Cohen]{mccloskey1989catastrophic}
McCloskey, Michael and Cohen, Neal~J.
\newblock Catastrophic interference in connectionist networks: The sequential
  learning problem.
\newblock In \emph{Psychology of learning and motivation}, volume~24, pp.\
  109--165. Elsevier, 1989.

\bibitem[Ollivier(2017)]{ollivier2017online}
Ollivier, Yann.
\newblock Online natural gradient as a kalman filter.
\newblock \emph{arXiv preprint arXiv:1703.00209}, 2017.

\bibitem[Polikar et~al.(2001)Polikar, Upda, Upda, and
  Honavar]{polikar2001learn++}
Polikar, Robi, Upda, Lalita, Upda, Satish~S, and Honavar, Vasant.
\newblock Learn++: An incremental learning algorithm for supervised neural
  networks.
\newblock \emph{IEEE transactions on systems, man, and cybernetics, part C
  (applications and reviews)}, 31\penalty0 (4):\penalty0 497--508, 2001.

\bibitem[Rhudy et~al.(2017)Rhudy, Salguero, and Holappa]{rhudy2017kalman}
Rhudy, Matthew~B, Salguero, Roger~A, and Holappa, Keaton.
\newblock A kalman filtering tutorial for undergraduate students.
\newblock \emph{International Journal of Computer Science \& Engineering Survey
  (IJCSES)}, 8:\penalty0 1--18, 2017.

\bibitem[Robins(1995)]{robins1995catastrophic}
Robins, Anthony.
\newblock Catastrophic forgetting, rehearsal and pseudorehearsal.
\newblock \emph{Connection Science}, 7\penalty0 (2):\penalty0 123--146, 1995.

\bibitem[Shin et~al.(2017)Shin, Lee, Kim, and Kim]{shin2017continual}
Shin, Hanul, Lee, Jung~Kwon, Kim, Jaehong, and Kim, Jiwon.
\newblock Continual learning with deep generative replay.
\newblock In \emph{Advances in Neural Information Processing Systems}, pp.\
  2990--2999, 2017.

\bibitem[Trebatick{\`y}(2005)]{trebaticky2005recurrent}
Trebatick{\`y}, Peter.
\newblock Recurrent neural network training with the extended kalman filter.
\newblock In \emph{IIT. SRC 2005: Student Research Conference}, pp.\ ~57, 2005.

\bibitem[Wo{\'z}niak et~al.(2014)Wo{\'z}niak, Gra{\~n}a, and
  Corchado]{wozniak2014survey}
Wo{\'z}niak, Micha{\l}, Gra{\~n}a, Manuel, and Corchado, Emilio.
\newblock A survey of multiple classifier systems as hybrid systems.
\newblock \emph{Information Fusion}, 16:\penalty0 3--17, 2014.

\bibitem[Wu \& Wang(2012)Wu and Wang]{wu2012extended}
Wu, Xuedong and Wang, Yaonan.
\newblock Extended and unscented kalman filtering based feedforward neural
  networks for time series prediction.
\newblock \emph{Applied Mathematical Modelling}, 36\penalty0 (3):\penalty0
  1123--1131, 2012.

\bibitem[Zenke et~al.(2017)Zenke, Poole, and Ganguli]{zenke2017continual}
Zenke, Friedemann, Poole, Ben, and Ganguli, Surya.
\newblock Continual learning through synaptic intelligence.
\newblock In \emph{Proceedings of the 34th International Conference on Machine
  Learning-Volume 70}, pp.\  3987--3995. JMLR. org, 2017.

\end{thebibliography}


\end{document}